\documentclass[accepted]{uai2025} 
                        
\usepackage[american]{babel}

\usepackage[numbers]{natbib} 
\usepackage{mathtools} 
\usepackage{booktabs} 
\usepackage{tikz} 

\usepackage{url}
\usepackage{caption}
\usepackage{algorithm,algorithmic}
\usepackage{amssymb}
\usepackage{amsthm}
\usepackage{xspace}
\usepackage{multirow}
\usepackage{tabularx}
\usepackage{enumitem}
\usepackage{subcaption}
\usepackage{caption}
\usepackage{colortbl}
\usepackage{arydshln}
\usepackage{wrapfig}
\usepackage{adjustbox}
\usepackage{multicol}

\theoremstyle{plain}
\newtheorem{theorem}{Theorem}[section]

\theoremstyle{definition}

\theoremstyle{remark}
\newtheorem{remark}[theorem]{Remark}

\newcommand{\beq}{\vspace{0mm}\begin{equation}}
\newcommand{\eeq}{\vspace{0mm}\end{equation}}
\newcommand{\beqs}{\vspace{0mm}\begin{eqnarray}}
\newcommand{\eeqs}{\vspace{0mm}\end{eqnarray}}
\newcommand{\barr}{\begin{array}}
\newcommand{\earr}{\end{array}}

\newcommand{\Amat}{{\bf A}}

\newcommand{\Imat}{{\bf I}}
\newcommand{\Jmat}{{\bf J}}

\newcommand{\ev}[0]{{\boldsymbol{e}}\xspace}

\newcommand{\mv}[0]{{\boldsymbol{m}}}

\newcommand{\xv}{\boldsymbol{x}}
\newcommand{\yv}{\boldsymbol{y}}

\newcommand{\deltav}[0]{{\boldsymbol{\delta}}\xspace}

\newcommand{\zeros}[0]{\boldsymbol{0}}

\usepackage{booktabs}
\usepackage{array}
\newcolumntype{L}[1]{>{\raggedright\let\newline\\\arraybackslash\hspace{0pt}}m{#1}}
\newcolumntype{C}[1]{>{\centering\let\newline\\\arraybackslash\hspace{0pt}}m{#1}}
\newcolumntype{R}[1]{>{\raggedleft\let\newline\\\arraybackslash\hspace{0pt}}m{#1}}

\newcommand{\R}{\mathbb{R}}

\newcommand{\argmax}{\operatorname{argmax}}

\newcommand{\name}{\textsc{Harvim}}

\definecolor{blue_ppt}{rgb}{0,0.47,0.97}

\title{
Beyond Invisibility: 
Learning Robust Visible Watermarks for Stronger Copyright Protection
}

\author[1]{\href{mailto:<liu3351@purdue.edu>?Subject=Your UAI 2025 paper}{Tianci Liu}{}}
\author[2]{\href{mailto:<tongyang@andrew.cmu.edu>?Subject=Your UAI 2025 paper}{Tong Yang}{}}
\author[3]{\href{mailto:<quan.zhang@broad.msu.edu>?Subject=Your UAI 2025 paper}{Quan Zhang}{}}
\author[4]{\href{mailto:<ql518@nyu.edu>?Subject=Your UAI 2025 paper}{Qi Lei}{}}

\affil[1]{%
Purdue University
}
\affil[2]{%
Carnegie Mellon University
}
\affil[3]{%
Michigan State University
}
\affil[4]{%
New York University
}

\begin{document}
\maketitle

\begin{abstract}

As AI advances, copyrighted content faces growing risk of unauthorized use, whether through model training or direct misuse. 
Building upon invisible adversarial perturbation, 
recent works developed copyright protections against specific AI techniques such as unauthorized personalization through DreamBooth that are misused. 
However, these methods offer only short-term security, as they require retraining whenever the underlying model architectures change.
To establish long-term protection aiming at better robustness, we go beyond invisible perturbation, and propose a universal approach that embeds \textit{visible} watermarks that are \textit{hard-to-remove} into images. 
Grounded in a new probabilistic and inverse problem-based formulation, our framework maximizes the discrepancy between the \textit{optimal} reconstruction and the original content. We develop an effective and efficient approximation algorithm to circumvent a intractable bi-level optimization.
Experimental results demonstrate superiority of our approach across diverse scenarios.

\end{abstract}

\section{Introduction}
\label{sec:intro}

Deep generative models (DGMs), such as diffusion models~\citep{sohl2015deep,ho2020denoising}, have shown remarkable success in various vision tasks, 
including text-to-image generation~\citep{rombach2022high}, 
image editing~\citep{choi2023custom,shi2024dragdiffusion}, and style transfer~\citep{kim2022diffusionclip}. 
Moreover, these models exhibit impressive personalization capabilities~\citep{gal2022image,ruiz2023dreambooth}.
For example, DreamBooth~\citep{ruiz2023dreambooth} fine-tunes diffusion models using a few representative reference images, enabling the generation of personalized images with high fidelity.
This capability significantly reduces the cost of AI-assisted personalized generation, paving the way for a wider range of AI-driven applications~\citep{yang2023diffusion}.

However, these advances also introduce new risks. 
Artists and photographers frequently share their works online for promotional purposes.
Yet, off-the-shelf AI tools enable malicious users to obtain unauthorized copies without purchasing rights or to directly plagiarize art styles by fine-tuning personalized models using these images~\citep{van2023anti,liu2024metacloak}. 
These threats greatly undermine the profits of art creators~\citep{shan2023glaze}.


Recent studies developed \textit{adversarial attacks} to defend against unauthorized use of AI tools~\citep{salman2023raising,shan2023glaze,liang2023mist,van2023anti}. 
These methods learn \textit{invisible perturbations }to {disrupt} the image generation process in DGMs like diffusion models. 
For example,~\cite{salman2023raising} push the latent codes of text-to-image diffusion models toward unrelated targets, 
and AdvDM~\citep{liang2023adversarial} minimizes the likelihood of perturbed images from diffusion models to degrade their performance on them. 
Poisoning attacks have also been used to trick fine-tuning based DreamBooth into learning false correlations, preventing it from capturing desired styles~\citep{van2023anti}, and MetaCloak~\citep{liu2024metacloak} incorporated meta-learning to attack an ensemble of diffusion models, improving the poisoning transferability.

Although effective on targeted models, these invisible attack-based solutions heavily rely on adversarial vulnerabilities, resulting in two key limitations. 
First, the \textit{adversarial attack-based} mechanism makes them fall short to generalize well to broader DGMs~\citep{huang2020metapoison,liu2024metacloak}.
Specifically, their performance on black-box DGMs is largely unpredictable~\citep{demontis2019adversarial}, and on white-box DGMs, they only provide \textit{short-term} protection: when facing new DGMs, the perturbation must also be updated or retrained~\citep{xue2024rethinking}. 
Second, their \textit{invisibility} inherently limits their strength from two aspects.
On one hand, invisible watermarks are prone to distortion and purification attacks~\citep{athalye2018synthesizing,liu2024metacloak,zhao2024can}. 
On the other, since these protections are designed to be invisible, they cannot prevent direct misuse such as scraping copyrighted content for commercial use without authorization.


In response, 
we propose a new paradigm for copyright protection.
Our approach revisits the visible watermark, a traditional tool for copyright protection. 
We demonstrate that \textit{visible} watermarks offer strong protection: 
with clear copyright information displayed, 
the image becomes largely unusable. 
Additionally, 
when a prominent visible watermark is present, 
AI tools like DreamBooth learn the watermark pattern due to their backdoor mechanism~\citep{rawat2022devil,pan2023trojan,chou2023backdoor}, resulting in unsatisfactory outputs.
Finally, our protection is agnostic to misuse:
unlike attack-based methods, visible watermarking does not target specific misuses or DGMs, thus providing a universal protection.

Another advantage of visible watermarking is its robustness to distortion attack, such as JPEG compression and Gaussian blur, that can easily compromise its invisible counterpart~\citep{athalye2018synthesizing,zhao2024can}. 
The existence of \textit{watermark removal} as \textit{targeted} attack on visible watermarking also poses a significant challenge~\citep{liu2021wdnet,liang2021visible,lugmayr2022repaint,liu2023aipo}. 
Since standard mechanism that adds visible watermarks in a consistent way can be bypassed by specialized attacks~\citep{dekel2017effectiveness}, and 
manually placing watermarks in appropriate areas~\citep{voyatzis1999use} can be labor-intensive and not scalable, we propose {\name}, which \textit{learns} a visible watermark that is hard to remove in an automated way.
\textit{To our best knowledge, this is the first learning-based visible watermark for copyright protection of human-created content in the AI era.
This new exploration is a key contributions.}

Formally, 
{\name} transforms watermark removal into an inpainting problem of reconstructing the watermarked area,
and learns a watermark to make the reconstruction harder. 
This entails a bi-level optimization. 
The lower-level optimization reconstructs the watermarked area, 
and the upper-level optimization adjusts the watermark to push the reconstruction away from the original image. 
Through this formulation, 
{\name} identifies a hard-to-reconstruct region of the image, 
usually containing rich visual details. 
Importantly, This region is an \textit{intrinsic} characteristic of the image, 
allowing {\name} to create watermarks that are \textit{inherently} hard to remove, regardless of the removal method used. 
\textit{The new hard-to-remove watermark formulation as a universal copyright protection is also a key contribution of this work.}


In execution, {\name} uses a pre-trained generative model as a \textit{prior} to guide lower-level optimization~\citep{bora2017compressed,asim2020invertible,ongie2020deep}. 
However, this generative prior makes the bi-level problem NP-hard~\citep{lei2019inverting,sinha2017review}, 
due to the complexity of how the watermark impacts the lower-level optimal solution involving a deep neural network (DNN).
Following prior work~\citep{huang2020metapoison,liu2024metacloak}, we use meta-learning~\citep{finn2017model} for an approximate solution, 
replacing the exact lower-level solution with one that takes $K$ gradient descent steps from the initial value~\citep{finn2017model}.
Expressing the gradients as functions of the watermark allows the approximation to be written as an explicit function of the watermark. 
Meta-learning requires $K$ to be small, usually leading to approximation errors~\citep{huang2020metapoison,geiping2020witches}.
Nonetheless, recent work showed that a special family of deep generative priors allows the lower-level optimization to be replaced by a series of subproblems, each solvable \textit{in a few steps}~\citep{liu2023aipo}. 
Built upon this, 
we derive a new, effective solution to learn watermark. 
\textit{This new bi-level solver is our third contribution.}

Our paper is organized as follows. 
Sec \ref{sec:method} discusses the {\name} formulation and its approximate solution.
Sec \ref{sec:experiment} evaluates {\name}'s performance on various image sets and tests its robustness against different watermark removers.
 Sec \ref{sec:related} reviews related works, and Sec \ref{sec:conclusion} conclude the paper.

\section{Proposed Method}
\label{sec:method}

Grounded in a probabilistic view, 
We propose {\name} to learn \textit{hard-to-remove} visible watermarks to protect copyrighted images from direct and AI-assisted misuse.

\subsection{Preliminary}

\textbf{Notations.}
As in previous works~\citep{ongie2020deep, whang21solve, liu2023aipo}, we represent (flattened) images as vectors denoted by lowercase boldface letters. Uppercase boldface letters mark matrices. 

\textbf{Inverse problems.} 
Given a corrupted observation $\yv \in \R^m$ of an unknown image $\xv_T \in \R^n$ ($m \leq n$), 
inverse problems aim to reconstruct clean $\xv_T$ assuming that $\yv$ is generated by
\begin{align}\label{eq:inv-prob}
   \yv = f(\xv_T) + \ev,
\end{align}
where $f(\cdot)$ is a known forward operator that corrupts $\xv_T$, and $\ev$ is a noise that has independently and identically distributed elements~\citep{bora2017compressed,ongie2020deep}.
Inverse problems, like compressed sensing and inpainting, are associated with a specific operator $f$.
For more background, see~\cite{ongie2020deep}. 
Our work focuses on image inpainting.

\textbf{Image inpainting.} 
This task aims to recover an image with masked content.
Formally, inpainting assumes that $\yv = \Amat \xv_T + \ev$, where $\Amat \in \R^{n \times n}$ is a diagonal matrix with binary entries indicating whether a pixel is observed or missing, 
and $\ev \sim \mathcal G(\zeros; \sigma^2 \Imat)$ is an isotropic Gaussian noise with known variance $\sigma^2$.

\textbf{Deep Generative Prior.} 
Inverse problems are generally under-determined, in the sense that
Eq \eqref{eq:inv-prob} admits infinitely many possible solutions.
To address this,  deep generative models (DGMs) pre-trained on large datasets can be used as \textit{priors} to assess the plausibility of reconstructions and help find the optimal one~\citep{ongie2020deep}. 
From a Bayesian perspective, this entails a \textit{maximum-a-posterior} (MAP) problem.
Let $G$ be a DGM prior.
We solve the inverse problem by finding
\begin{align}
    \xv^*
    &=
    \argmax
    \nolimits_{\xv} \log p_G(\xv \mid \yv; \lambda) \\
    &= 
    \argmax
    \nolimits_{\xv} \log p_{e}(\yv - f(\xv)) + \lambda \log p_G(\xv), \notag
\end{align}
$\log p_e (\cdot)$ and $\log p_G (\cdot)$ represent the log-likelihood of noise $\ev$ and image $\xv$, respectively. 
The hyperparameter $\lambda > 0$ controls the weight of the prior $G$, acting as a regularizer~\citep{whang21solve}.

\textbf{Copyrighted Image Protection.} 
The advance of DGMs also enables unauthorized use of copyrighted content. 
For instance, DreamBooth~\citep{ruiz2023dreambooth} allows text-to-image diffusion models~\citep{rombach2022high} to generate personalized images. 
However, by fine-tuning on a few of an artist's work, it can mimic and plagiarize their style~\citep{van2023anti}.
This has raised significant concerns about copyright protection~\citep{shan2023glaze}. 
To counter this, 
recent works~\citep{liang2023mist,van2023anti} proposed \textit{targeted} attacks on DGMs like DreamBooth being misused. 
Conceptually, 
given a misused DGM $G$ with training loss $\ell(G; \xv)$ for any $\xv$, these works protected copyrighted image $\xv_T$ by learning an invisible perturbation $\deltav$ via $\max\nolimits_{\deltav: \| \deltav \|_\infty < \varepsilon} \ell(G; \xv + \deltav)$ to degrade $G$'s performance on $\xv+\deltav$, where $\varepsilon$ limits pixel-level perturbation. 
This defines an adversarial attack on $G$. 
When $G$ is inaccessible, $\delta$ is learned by attacking (an ensemble of) open-source surrogate models~\citep{liu2024metacloak}.

\subsection{{\name}: Towards a Universal Protection by Visible Watermarking}


As outlined before, 
although attack-based safeguards can effectively address targeted misuse, they have key weakness. 
First, the attack-based formulation limits their applicability in \textit{untargeted} scenarios.
Specifically, 
their performance on black-box AI is largely unpredictable due to the nature of the attack~\citep{demontis2019adversarial,liu2024metacloak},
and in white-box settings, they provide only \textit{short-term} protection, in the sense that 
new personalization techniques may render current safeguards (e.g., those against DreamBooth) ineffective~\citep{liu2024metacloak,xue2024rethinking}. 
In addition, the \textit{invisible} nature of existing protections also poses two inherent limitations.
First, these protections are prone to distortion or purification attack~\citep{athalye2018synthesizing,liu2024metacloak,zhao2024can}. 
Second, the \textit{invisibility} offers no protection against \textit{direct misuse}.
We refer to a misuse as direct if it does not involve AI, but rather unauthorized use such as piracy. 
For instance, users may scrape copyrighted images for commercial purposes without purchasing rights, undermining creators' profits. Such misuse doesn't involve AI tools and cannot be addressed by existing attack-based methods.
Consequently, existing safeguards often provide unsatisfactory protection in execution~\citep{liu2024metacloak,zhao2024can}. 
Hence, a more general formulation for protection is needed. 

\begin{figure*}[htb!]
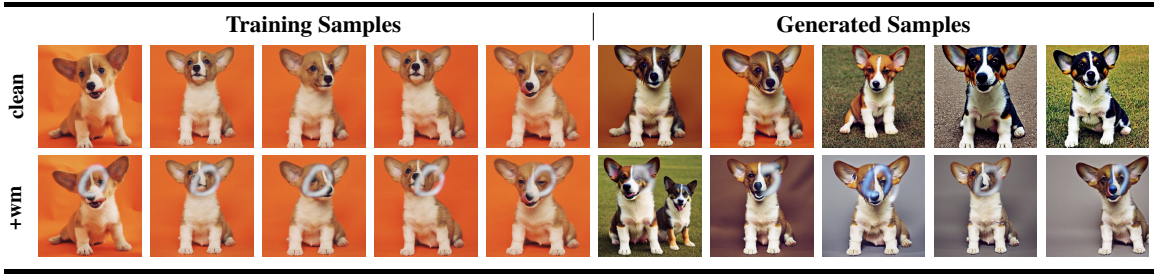

\centering
\resizebox{0.9\textwidth}{!}{
\renewcommand{\tabcolsep}{2pt}
\def\figwidth{0.09\linewidth}%
\newcommand{\authornote}[1]{
\adjustbox{rotate=90}{\parbox{\figwidth}{\footnotesize \bf \centering #1}}
}

\begin{tabular}{*{12}{c}}

\toprule[0.4ex]
& \multicolumn{5}{c}{\bf Training Samples}  \vline
& \multicolumn{5}{c}{\bf Generated Samples} \\
\noalign{\vskip 0.5ex}

\authornote{clean}     &  
\loadimage{figures/dbooth/train}{width=\figwidth}{0}{5}
&
\loadimage{figures/dbooth/gen}{width=\figwidth}{0}{5}\\

\authornote{+wm}     &  
\loadimage{figures/dbooth/train_wm}{width=\figwidth}{0}{5}
&
\loadimage{figures/dbooth/gen_wm}{width=\figwidth}{0}{5}\\
\bottomrule[0.4ex]

\end{tabular}
}

\caption{
Visible Watermarking can provide strong protection: 
DreamBooth trained on watermarked (``+wm'') images learn watermark patterns as well.
Examples and implementations are from \cite{von2022diffusers}.
}
\label{fig:dbooth}
\end{figure*}
\begin{figure}[htb!]
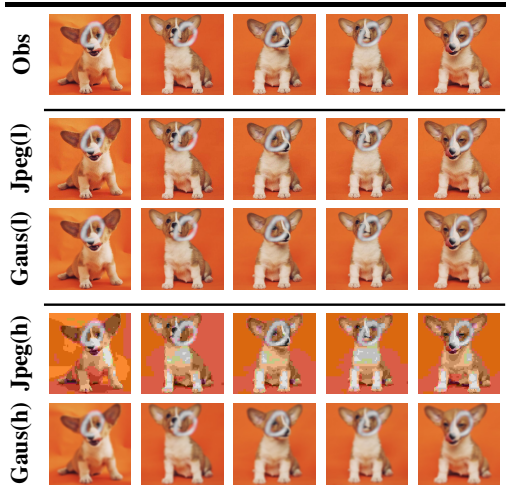

\centering
\renewcommand{\tabcolsep}{2pt}
\def\figwidth{0.13\linewidth}%
\newcommand{\authornote}[1]{
\adjustbox{rotate=90}{\parbox{\figwidth}{ \bf \small \centering #1}}
}

\begin{tabular}{cccccc}

\toprule[0.4ex]


\authornote{Obs}     &  
\loadimage{figures/distortion/dog_wm}{width=\figwidth}{0}{5} \\

\cmidrule[0.2ex]{2-6}

\authornote{Jpeg(l)}     &  
\loadimage{figures/distortion/jpeg_high}{width=\figwidth}{0}{5} \\

\authornote{Gaus(l)}     &  
\loadimage{figures/distortion/gaus_high}{width=\figwidth}{0}{5} \\

\cmidrule[0.2ex]{2-6}

\authornote{Jpeg(h)}     &  
\loadimage{figures/distortion/jpeg_low}{width=\figwidth}{0}{5} \\

\authornote{Gaus(h)}     &  
\loadimage{figures/distortion/gaus_low}{width=\figwidth}{0}{5} \\

\bottomrule[0.4ex]

\end{tabular}
\caption{
Visible watermarks remain resilient to strong distortion attacks JPEG compression and Gaussian blur, at low- (top) and high-intensity (bottom) levels. 
}
\label{fig:harvim-distort}
\end{figure}

In light of these limitations, 
we resort to visible watermarking for stronger protection. 
First, visible watermarks render protected images largely unusable in {direct} use. In AI-involved misuse scenarios, when a prominent watermark presents, AI such as personalization with DreamBooth will also be affected due to the backdoor mechanism~\citep{rawat2022devil,pan2023trojan,chou2023backdoor}. 
As shown in Fig \ref{fig:dbooth}, DreamBooth learns watermark patterns from watermarked training images, leading to unusable outputs. 
Notably, adding visible watermarks requires no prior domain knowledge of misuse scenarios or mechanisms. Thus, it provides a broader protection. 
In addition, 
visible watermarking are much more robust to distortion attacks. 
In Fig \ref{fig:harvim-distort} we applied JPEG compression~\citep{dziugaite2016study,aydemir2018effects} and Gaussian blur~\citep{zhao2020blurring} at varied intensities to distort watermarked images, and observed that the watermarks remain readable even when the images are greatly destroyed.

Our finding indicates that visible watermark offers an excellent level of robustness against standard transformation attack, and pave the way for more reliable copyright protection than existing attempts. 
Nonetheless, conventional watermarks are typically added in a consistent manner to the images, which offers limited resistance against more targeted watermark removal attack~\citep{dekel2017effectiveness,liang2021visible,sun2023denet}. 
While manual or rule-based watermark placement can provide some protection~\citep{kankanhalli1999adaptive}, it requires significant human effort and lacks scalability.
To address this, 
we propose an \textit{automated} solution by \textit{learning} a visible watermark that is resistant to remove. 
We refer to our approach as \textit{\underline{ha}rd-to-\underline{r}emove \underline{vi}sible water\underline{m}ark} ({\name}) and provide details below.

\ExplSyntaxOn
\cs_set:Npn \loadimage #1#2#3#4 {
    \seq_clear:N \l_tmpa_seq
    \int_step_inline:nnn {#3} {#3 + #4 - 1} {
        \seq_put_right:Nx \l_tmpa_seq {
            \exp_not:N \includegraphics
                [\exp_not:n {#2}]
                {#1/##1.png}
        }
    }
    \seq_use:Nn \l_tmpa_seq {&}
}

\cs_set:Npn \calctotalwidth #1#2 {
    \fp_eval:n {
        (#2) * (#1) + (#2 - 1) * (\tabcolsep)
    } pt
}
\ExplSyntaxOff

\ExplSyntaxOn
\cs_set:Npn \loadimage #1#2#3#4 {
    \seq_clear:N \l_tmpa_seq
    \int_step_inline:nnn {#3} {#3 + #4 - 1} {
        \seq_put_right:Nx \l_tmpa_seq {
            \exp_not:N \includegraphics
                [\exp_not:n {#2}]
                {#1/##1.png}
        }
    }
    \seq_use:Nn \l_tmpa_seq {&}
}

\cs_set:Npn \calctotalwidth #1#2 {
    \fp_eval:n {
        (#2) * (#1) + (#2 - 1) * (\tabcolsep)
    } pt
}
\ExplSyntaxOff

\subsection{Formal Formulation of {\name}}


We formulate the proposed {\name} as an optimization problem. 
To this end, we define a {watermark} $\mv \in \R^n$ as an image with the same dimensions\footnote{The background is also part of the image.} as the copyrighted image $\xv_T$. 
Then, watermark removal can be formulated as an inverse problem~\citep{ongie2020deep}, 
where the watermarked observation is%
\footnote{We write the observation $\yv$ (or reconstruction $\xv^*$) as a function of $\mv$ (and hyperparameter $\lambda$) to highlight the dependence. 
} 
$\yv(\mv) = \Amat_m \xv_T + \ev$. 
Similar to inpainting, $\Amat_m$ is a diagonal matrix where entries indicate if a pixel is watermarked. Treating the {watermarked} area as missing, inpainting serves as a surrogate for visible watermark removal~\citep{huang2004attacking}.

Built upon this formulation, {\name} seeks an $\mv$ that makes $\xv_T$ \textit{hard to reconstruct} from observation $\yv(\mv)$.
The \textit{reconstruction hardness} is measured by a {similarity score} $s(\xv^*(\mv), \xv_T)$ between the optimal reconstruction $\xv^*(\mv)$ from $\yv (\mv)$ to the ground truth $\xv_T$.

\textbf{Watermarking constraints.}
When learning $\mv$ for copyrighted image protection, two standard \textit{readability} constraints must be met
\citep{mohanty1999dual,kankanhalli1999adaptive}.
First, 
\textbf{image readability}
requires that the watermarked observation's readability must remain. 
Otherwise, while an excessive watermark occupying the entire image can make it unrecoverable, 
audience will also fail to recognize the image content, which could negatively compromise the creator's financial gains and public visibility.
This constraint is solved by adding a regularization term $\mathcal R (\mv)$ to penalize the size of watermark.
Second, \textbf{watermark readability}
requires that the watermark itself should convey clear copyright information, such as the creator's logo or name.
To satisfy this constraint,
we use a small pre-trained generative model to control $\mv$, as detailed in Appendix \ref{app:wm-details} due to page limit.

Put together, 
{\name} learns $\mv$ to watermark image $\xv_T$ by solving a bi-level optimization problem
\begin{align}\label{eq:opt-wm}
    &\mv^* 
    = 
    \min\nolimits_{\mv} 
    s(\xv^*(\mv), \xv_T) + \mathcal{R}(\mv), \\
    &\text{s.t.}\ 
    \xv^*(\mv) 
    = 
    \argmax\nolimits_{\xv} \log p_G (\xv \mid \yv(\mv); \lambda). \notag
\end{align}


\begin{remark}
We want to emphasize that Eq \eqref{eq:opt-wm} provides a general framework for image protection for two key reasons. 
First, the concept of \textit{hard-to-reconstruct region} underlying {\name} reflects an intrinsic characteristic of an image, rather than a property specific to any particular prior $G$. 
Second, Eq \eqref{eq:opt-wm} is not limited to any specific choice of $G$. 
The next section presents an implementation, 
but {\name} by definition can incorporate any generative prior $G$ capable of modeling the real image distribution.
\end{remark}

\subsection{An Approximate Solution for {\name}}

The bi-level optimization Eq \eqref{eq:opt-wm} is non-trivial to solve, with difficulties lying in two folds. 
First, inpainting requires matrix $\Amat_m$ containing binary entries, which cannot be optimized by gradient-based method. 
Second, its feasible set, as specified by the lower-level optimization that involves some deep neural network $G$, is NP-hard to identify~\citep{sinha2017review}.
Therefore, further approximations are needed.

Mathematically, the first challenge arises from that $\mv$'s gradient is undefined due to the discrete nature of $\Amat_m$.
To address this issue, we construct a differentiable approximation for it based on continuous-valued learnable watermark $\mv$.  
Specifically, given $\mv \in \R^n$, denote the sigmoid function by $\text{sig}: \R \rightarrow \R$, we define 
\begin{align}\label{eq:inpaint-mask}
    \Amat_m = \text{diag}\left(\text{sig}\left(\frac{m_1 - \alpha}{\beta}\right), \dots, \text{sig}\left(\frac{m_n - \alpha}{\beta}\right)  \right),
\end{align}
where $\alpha, \beta$ are hyperparameters such that
$\text{sig}((m_i - \alpha) / \beta) \approx 1$ when $m_i$ lies within the watermark area, and $0$ otherwise. 
Additional implementation details are provided in Appendix \ref{app:wm-details}.

The differentiable $\mv$ can be optimized with gradient
\begin{align}\label{eq:bilevel-grad}
    &\nabla_{\mv} \left(s(\xv^*(\mv), \xv_T) + \mathcal{R}(\mv) \right) \\
    =& 
    \nabla_{\xv^*} s(\xv^*(\mv), \xv_T)^\top \Jmat_{\mv}(\xv^*(\mv)) + \nabla_\mv \mathcal R(\mv), \notag
\end{align}
where the second line holds from the chain rule, and $\Jmat_{\mv}(\xv^*(\mv))$ denotes the Jacobian of $\xv^*$ with respect to $\mv$. 
Unfortunately, this Jacobian is intractable due to the unknown form of $\xv^*(\mv)$, making the problem remain unsolved. 
We resort to meta-learning for an approximate solution~\citep{huang2020metapoison} by replacing the exact $\xv^*(\mv)$ with an approximate solution $\Tilde{\xv}(\mv)$ that is computed from $K$-step gradient descent~\citep{finn2017model}. 
By treating $\nabla_{\xv} \log p_G( \xv \mid \yv(\mv); \lambda)$
as a function of $\mv$, $\Tilde{\xv}(\mv)$ can be expressed as an explicit function of $\mv$, making approximation Eq \eqref{eq:bilevel-grad} viable. 



In practice, however, meta-learning requires small $K$ to maintain affordable computational cost, and 1 or 2 is often used~\citep{huang2020metapoison}. Such a small value often results in highly inaccurate approximation~\citep{geiping2020witches}, 
{Critically, when the approximation fails to reflect the faithful progress made by current $\mv$, 
the upper-level optimization will be misled as well, resulting in poor or failed solutions~\citep{ghadimi2018approximation,franceschi2018bilevel}.
}

\textbf{Idea.}
While this difficulty cannot be resolved in general, 
for inverse problems solvers that use normalizing flows~\citep{papamakarios2021normalize} as generative priors, it can be largely alleviated.
In specific, denote $\xv^*(\yv; \lambda) = \argmax\nolimits_{\xv} \log p_G(\xv \mid \yv; \lambda)$,~\cite{liu2023aipo} showed that under regular conditions, $\log p_G(\xv \mid \yv; \lambda')$ is locally convex at $\xv^*(\yv; \lambda)$ when $\lambda'$ is close enough to $\lambda$. Therefore, using $\xv^*(\yv; \lambda)$ as an initial value, $\xv^*(\yv; \lambda')$ by nature can be obtained within \textit{a few} gradient descent steps.
Motivated by this, 
we expect $\log p_G(\xv \mid \yv; \lambda)$ to preserve a local convexity around $\xv^*(\yv'; \lambda')$ if $\yv'$ is close to $\yv$ and $\lambda'$ is close to $\lambda$.
Built upon this, 
we optimize $\mv$ along with $\Tilde{\xv}(\mv)$ and $\lambda$ as in~\cite{liu2023aipo} together in an iterative way. 


\textbf{Solution.}
Our solution starts with a randomly initialized watermark $\mv_0$, hyperparameter $\lambda_0 = 0$, and an approximate solution $\Tilde{\xv}(\mv_0; \lambda_0)$ solved by gradient descent.
Here the approximate solution $\Tilde{\xv}$ is expressed as a function of both watermark $\mv$ and $\lambda$.
In each round $t$, 
we first update hyperparameter $\lambda_t$ by taking a small step towards the final $\lambda$.
Next, given current $\mv_{t-1}$ and $\lambda_t$, we solve $\Tilde{\xv}_t(\mv_{t-1}; \lambda_{t})$ by taking $K$ gradient descent steps from the last round solution $\Tilde{\xv}_{t-1}$.
Finally, we update $\mv_t$ by {unrolling} updates on $\Tilde{\xv}_t(\mv_{t-1}; \lambda_{t})$ as a function of $\mv_{t-1}$ and take
\begin{align}\label{eq:wm-update}
\mv_t = \mv_{t-1} - \nabla_\mv \left( s(\underbrace{\Tilde{\xv}_t(\mv_{t-1}; \lambda_{t})}_{\text{func. of $\mv_{t-1}$}}, \xv_T) + \mathcal R(\mv_{t-1}) \right).
\end{align}
We repeat the following steps until $\lambda_t$ reaches the pre-specified value $\lambda$. 
The solution is outlined in Algo \ref{alg:main}.

\begin{algorithm}[!t]
\caption{{\name} algorithm}
\label{alg:main}

\begin{algorithmic}[1] 
    \STATE \textbf{Input:}  
    copyrighted image $\xv_T$,
    $\lambda>0$ and its update steps $T > 0$,
    random noise variance $\sigma^2 > 0$,
    generative prior $G:\R^n\to \R^n$,
    inpainting mask hyperparameters $\alpha, \beta$ (Eq \eqref{eq:inpaint-mask}),
    unrolled steps $K$
    
    \STATE \textbf{Initialize:}
    $\lambda_0=0$, 
    randomly initialize $\mv_0$ and inpainting mask $\Amat_{m, 0}$ based on Eq \eqref{eq:inpaint-mask},
    watermarked image $\yv_0 = \Amat_{m, 0} \xv_T + \ev$ where $\ev \sim \mathcal G(\zeros, \sigma^2 \Imat)$
    \STATE 
    Ignoring dependency on $\mv_0$, 
    find the MLE solution $\Tilde{\xv}_0 = \Tilde{\xv}(\mv_0, \lambda_0)$ for $\yv_0$~\citep{liu2023aipo}
    \FOR{$t=1, \dots, T$} 
        \STATE
        Treat $\yv_{t-1}(\mv_{t-1}) = \Amat_{m, t-1} \xv_T + \ev, \ev \sim \mathcal G(\zeros, \sigma^2 \Imat)$ as a function of $\mv_{t-1}$
        \STATE 
        $\lambda_t = \lambda_{t-1} + \frac{\lambda}{T}$
        \STATE 
        $\Tilde{\xv}_t = \Tilde{\xv}_{t-1}$
        \FOR{$k=1, \dots, K$}
            \STATE
            $\Tilde{\xv}_t = \Tilde{\xv}_t + \nabla_{\xv} \log p(\xv \mid \yv_{t-1}(\mv_{t-1}); \lambda_t)$
        \ENDFOR
        \STATE
        Denote current solution as $\Tilde{\xv}_t(\mv_{t-1}, \lambda_t)$
        \STATE\label{step:update_m}
        \textcolor{black}{
        Update $\mv_t$ based on Eq \eqref{eq:wm-update}
        }
    \ENDFOR 
    \RETURN Learned $\mv_t$
\end{algorithmic}
\end{algorithm}

\section{Experiments}
\label{sec:experiment}

In this section, we evaluate the performance of {\name} in learning various types of watermarks across diverse image distributions. Importantly, 
{\name} employs a simpler $G$ to acquire information about \textit{reconstruction hardness} for guiding watermark optimization, and learned watermarks are capable of resisting more advanced watermark removal techniques. These results confirmed the versatility of {\name}. 

\subsection{Experiment Setup}

\textbf{{\name} Setup.}
Following~\citet{whang21solve,liu2023aipo}, we use representative normalizing flow model RealNVP~\citep{dinh2016density} as pre-trained on CelebA~\citep{liu2015faceattributes} as a reliable generative prior $G$ \citep{liu2020empirical}. 
More training details can be found in~\citet{whang21solve}. 
We adopt peak-signal-to-ratio (PSNR) to measure similarity between reconstruction $\Tilde{\xv}$ and ground truth $\xv_T$ that {\name} seeks to minimize in Eq \eqref{eq:opt-wm}.

\textbf{Learnable Watermarks.}
We consider two families of learnable watermarks for empirical study. 
The \textbf{logo-styled} watermarks are simulated by MNIST digits~\citep{lecun1998gradient}, and we use all digits 0-9.
The \textbf{initial-styled} watermarks, on the other hand, are constructed from handwritten English letters~\citep{cohen2017emnist}, 
and we choose two randomly selected initials, ``NJ'' and ``OS''. 
All watermark generators are implemented by lightweight variational auto-encoder (VAE,~\citet{kingma2013auto}) using fully-connected layers and can be trained with CPU only. 
We provide more details in Appendix \ref{app:wm-details}.

\textbf{Image Datasets.}
We consider three image sets to protect.
The \textbf{In-distribution} set is a validation subset of CelebA whereon $G$ was trained. 
This dataset helps understand the scenario where $G$ can be maintained by the copyright owners. 
For further evaluations of {\name} in scenarios where copyrighted images are not allowed to be used for training $G$, 
we consider two \textbf{out-of-distribution} sets:
a validation subset of ImageNet~\citep{deng2009imagenet}, and 10 manually selected Cartoon images. 
Due to budget constraints, on CelebA and ImageNet we randomly choose 100 images respectively, see Appendix \ref{app:data} for more details.

\textbf{Watermark Removal Methods.}
After constructing \textit{hard-to-remove} watermark $\mv$, 
we conduct two classes of watermark removal methods. 
The first \textit{worst-case} class have access to the ground truth location of watermarks (i.e., exact $\Amat_m$ is assumed known) and remove them by solving inverse problems. To this end, \textbf{Flow-R} uses the same flow-based model $G$ to solve the inpainting task with random initialized $\xv$~\citep{liu2023aipo}, and \textbf{RePaint} is a representative diffusion model-based inpainting method~\citep{lugmayr2022repaint}. 
The second \textit{Blind-case} class contains \textbf{SLBR} \citep{liang2021visible} and \textbf{DeNet} \citep{sun2023denet}, which are blind watermark removal models that are pretrained on diverse images and watermarks to \textit{locate-and-remove} watermark in an end-to-end manner. 
Notably, \textit{{\name} is not optimized for any of these methods.}


\textbf{Evaluation Metrics.}
We evaluate the performance of {\name} based on the reconstruction quality of $\Tilde{\xv}(\mv_0)$ and $\Tilde{\xv}(\mv_T)$, where $\mv_0$ and $\mv_T$ denotes the initial and learned watermarks respectively. Following the literature~\citep{liang2021visible,lugmayr2022repaint}, the reconstruction quality is measured by peak-signal-to-ratio (PSNR),  structural similarity (SSIM), and learned perceptual image patch similarity (LPIPS) \citep{zhang2018unreasonable}.
As will be detailed shortly, we manipulate the three metrics to make sure that 
higher indicates better reconstruction, and thus weaker copyright protection.

\subsection{Quantitative Evaluation of {\name}}

When conducting watermark removal, 
We noted all of the four methods suffered from notable performance degradation in challenging scenarios. 
As an extreme case, SLBR and DeNet failed to recognize watermarks, and produced reconstruction nearly identical to the watermarked observation, as shown in Fig \ref{fig:main}.
Consequently, a direct comparison of PSNR and other metrics may fail to correctly measures the effectiveness of {\name}:
when a reconstruction $\Tilde{\xv}$ is identical to the observation $\yv$, metric $\text{PSNR}(\Tilde{\xv}, \xv_T) = \text{PSNR}(\yv, \xv_T)$ 
in essence quantifies \textit{how much watermark $\mv$ distorts the image}, other than \textit{how difficult it is to be removed}. 

To avoid this misleading evaluation, we check \textit{to what extent a reconstructed image is better than the watermarked observation} by computing how much PSNR or SSIM from a reconstruction to the ground truth is higher than from the observation. 
Specifically, we defined 
$v_\text{PSNR}(\xv) \triangleq \text{PSNR}(\Tilde{\xv}, \xv_T) - \text{PSNR}(\yv, \xv_T)$
as a measure of how good reconstruction $\Tilde{\xv}$ is in terms of PSNR, the measures of SSIM and LPIPS are defined similarly\footnote{As lower LPIPS implies higher similarity, we flip its subtraction order to make larger $v_\text{LPIPS}$ indicate better reconstruction. }. 
We report these results (mean$\pm$se) in Tab \ref{tab:quant}. 
Due to page limit, we defer the results from SLBR that failed on our watermarks to App \ref{app:results}. Original metrics are also reported in Tab \ref{tab:raw} in App \ref{app:results} for more comprehensive evaluation.

From Tab \ref{tab:quant}, {\name} successfully learned watermarks resisting both flow- and diffusion-based worst-case methods,
Flow-R and RePaint, in all cases. 
Blind-case methods failed to identify added watermarks,
possibly due to the substantial style and semantic difference between our learned watermarks and their pre-trained data. This highlights the limitation of blind-case methods.

When comparing the defense performance against the two worst-case methods, 
{\name} exhibited better performance on Flow-R than on RePaint. 
We hypothesize that this superior performance can be attributed to the fact that {\name} and Flow-R share the same generative prior $G$ and employ a similar \textit{maximum-a-posteriori} Bayesian optimization framework.
In contrast, images reconstructed by RePaint undergo a significantly different optimization process. 
Conceptually, this distinction is similar to attacking a \textit{gray-box} model versus \textit{black-box} model~\citep{papernot2017practical}.
Furthermore, {\name} shows strong transferability in both scenarios, which are challenging for traditional adversarial attacks~\citep{demontis2019adversarial}.
We attribute this success to the fact that {\name}'s target, the \textit{hard-to-reconstruct region} of image $\xv_T$, 
is an intrinsic characteristic of the real $\xv_T$. Consequently, any generative model pre-trained on real images will inherently reflect this property.
As a result, {\name} offers a general protection. 
In contrast, previous adversarial attack-based protections targeted on models-specific shortcuts that are not shared across different models, often resulting in unsatisfactory transferability~\citep{huang2020metapoison}.


As further evidence, although the generative prior $G$ was trained on CelebA, 
{\name} still offers comparable defense performance on out-of-distribution ImageNet and Cartoon datasets. 
As pointed in previous studies~\citep{asim2020invertible, whang21solve}, 
Flow as a generative prior provides a certain degree of generalizability across different image distributions for measuring the likelihood of an image. Our results further demonstrate that this flexibility can be leveraged to identify the \textit{hard-to-reconstruct region} in out-of-distribution images as well. 


\begin{table*}[htb!]
\definecolor{verylightgray}{gray}{0.95}
\setlength{\tabcolsep}{4pt}  

\centering
\caption{
Worse-case performance of {\name} when removers know the exact position of watermarks. 
The performance is evaluated based on to what extent the watermark removing performance is better than the observation in terms of PSNR, SSIM, and LPIPS respectively. Lower indicates worse reconstruction quality, thus stronger protection.  
}
\label{tab:quant}
\resizebox{0.75\linewidth}{!}{%
\begin{tabular}{
c >{\bfseries}r 
cc>{\columncolor{verylightgray}}c 
cc>{\columncolor{verylightgray}}c 
cc>{\columncolor{verylightgray}}c
}
\toprule[0.4ex]
\multicolumn{11}{c}{\bf CelebA} \\
\cmidrule[0.15ex]{2-11}

& & \multicolumn{3}{c}{\bf PSNR} & \multicolumn{3}{c}{\bf SSIM $\times$ 100} & \multicolumn{3}{c}{\bf LPIPS $\times$ 100} \\
\cmidrule[0.1ex]{3-11}
& & \bf Random & \bf {\name} & \bf Imp ($\uparrow$) & \bf Random & \bf {\name} & \bf Imp ($\uparrow$) & \bf Random & \bf {\name} & \bf Imp ($\uparrow$) \\
\cmidrule(lr){3-5} \cmidrule(lr){6-8} \cmidrule(lr){9-11}

\multirow{2}{*}{\rotatebox{90}{\bf \small DIG}} 
& Flow-R & $13.02_{\pm 0.37}$ & $7.57_{\pm 0.66}$ & 5.44 & $6.71_{\pm 0.24}$ & $3.82_{\pm 0.40}$ & 2.89 & $3.22_{\pm 0.27}$ & $2.21_{\pm 0.24}$ & 1.01 \\
& RePaint & $13.45_{\pm 0.41}$ & $11.57_{\pm 0.46}$ & 1.89 & $9.63_{\pm 0.25}$ & $8.62_{\pm 0.33}$ & 1.01 & $5.96_{\pm 0.27}$ & $5.74_{\pm 0.21}$ & 0.22 \\

\noalign{\vskip 0.3ex}\cdashline{3-11}\noalign{\vskip 0.3ex}
\multirow{2}{*}{\rotatebox{90}{\bf \small NJ}}
& Flow-R & $9.31_{\pm 0.34}$ & $7.33_{\pm 0.36}$ & 1.98 & $11.40_{\pm 0.58}$ & $9.34_{\pm 0.61}$ & 2.06 & $7.53_{\pm 0.39}$ & $6.44_{\pm 0.44}$ & 1.08 \\
& RePaint & $10.25_{\pm 0.34}$ & $9.99_{\pm 0.38}$ & 0.26 & $14.08_{\pm 0.60}$ & $13.07_{\pm 0.67}$ & 1.01 & $9.36_{\pm 0.34}$ & $8.93_{\pm 0.41}$ & 0.43 \\

\noalign{\vskip 0.3ex}\cdashline{3-11}\noalign{\vskip 0.3ex}
\multirow{2}{*}{\rotatebox{90}{\bf \small OS}}
& Flow-R & $9.57_{\pm 0.33}$ & $7.21_{\pm 0.31}$ & 2.36 & $11.43_{\pm 0.62}$ & $8.01_{\pm 0.53}$ & 3.42 & $7.90_{\pm 0.41}$ & $7.02_{\pm 0.39}$ & 0.88 \\
& RePaint & $10.39_{\pm 0.36}$ & $9.44_{\pm 0.37}$ & 0.96 & $15.68_{\pm 0.64}$ & $12.86_{\pm 0.62}$ & 2.81 & $10.62_{\pm 0.43}$ & $10.09_{\pm 0.41}$ & 0.53 \\

\cmidrule[0.15ex]{2-11}

\multicolumn{11}{c}{\bf ImageNet} \\
\cmidrule[0.15ex]{2-11}

& & \multicolumn{3}{c}{\bf PSNR} & \multicolumn{3}{c}{\bf SSIM $\times$ 100} & \multicolumn{3}{c}{\bf LPIPS $\times$ 100} \\
\cmidrule[0.1ex]{3-11}
& & \bf Random & \bf {\name} & \bf Imp ($\uparrow$) & \bf Random & \bf {\name} & \bf Imp ($\uparrow$) & \bf Random & \bf {\name} & \bf Imp ($\uparrow$) \\
\cmidrule(lr){3-5} \cmidrule(lr){6-8} \cmidrule(lr){9-11}

\multirow{2}{*}{\rotatebox{90}{\bf \small DIG}}
& Flow-R & $9.61_{\pm 0.45}$ & $6.09_{\pm 0.62}$ & 3.52 & $4.49_{\pm 0.35}$ & $2.72_{\pm 0.45}$ & 1.77 & $3.22_{\pm 0.27}$ & $2.21_{\pm 0.24}$ & 1.01 \\
& RePaint & $9.61_{\pm 0.48}$ & $8.56_{\pm 0.53}$ & 1.05 & $8.80_{\pm 0.39}$ & $8.40_{\pm 0.44}$ & 0.41 & $5.96_{\pm 0.27}$ & $5.74_{\pm 0.21}$ & 0.22 \\

\noalign{\vskip 0.3ex}\cdashline{3-11}\noalign{\vskip 0.3ex}
\multirow{2}{*}{\rotatebox{90}{\bf \small NJ}}
& Flow-R & $8.46_{\pm 0.37}$ & $7.21_{\pm 0.42}$ & 1.25 & $5.12_{\pm 0.73}$ & $3.42_{\pm 0.75}$ & 1.70 & $7.53_{\pm 0.39}$ & $6.44_{\pm 0.44}$ & 1.08 \\
& RePaint & $7.20_{\pm 0.31}$ & $6.69_{\pm 0.40}$ & 0.50 & $3.76_{\pm 0.73}$ & $2.79_{\pm 0.82}$ & 0.96 & $9.36_{\pm 0.34}$ & $8.93_{\pm 0.41}$ & 0.43 \\

\noalign{\vskip 0.3ex}\cdashline{3-11}\noalign{\vskip 0.3ex}
\multirow{2}{*}{\rotatebox{90}{\bf \small OS}}
& Flow-R & $8.58_{\pm 0.34}$ & $7.18_{\pm 0.39}$ & 1.40 & $4.28_{\pm 0.69}$ & $1.82_{\pm 0.69}$ & 2.46 & $7.90_{\pm 0.41}$ & $7.02_{\pm 0.39}$ & 0.88 \\
& RePaint & $7.44_{\pm 0.33}$ & $6.60_{\pm 0.38}$ & 0.84 & $4.03_{\pm 0.68}$ & $2.03_{\pm 0.77}$ & 2.00 & $10.62_{\pm 0.43}$ & $10.09_{\pm 0.41}$ & 0.53 \\

\cmidrule[0.15ex]{2-11}
\multicolumn{11}{c}{\bf Cartoon} \\
\cmidrule[0.15ex]{2-11}

& & \multicolumn{3}{c}{\bf PSNR} & \multicolumn{3}{c}{\bf SSIM $\times$ 100} & \multicolumn{3}{c}{\bf LPIPS $\times$ 100} \\
\cmidrule[0.1ex]{3-11}
& & \bf Random & \bf {\name} & \bf Imp ($\uparrow$) & \bf Random & \bf {\name} & \bf Imp ($\uparrow$) & \bf Random & \bf {\name} & \bf Imp ($\uparrow$) \\
\cmidrule(lr){3-5} \cmidrule(lr){6-8} \cmidrule(lr){9-11}

\multirow{2}{*}{\rotatebox{90}{\bf \small DIG}}
& Flow-R & $6.53_{\pm 1.18}$ & $-0.86_{\pm 1.74}$ & 7.39 & $2.55_{\pm 0.77}$ & $-0.88_{\pm 1.25}$ & 3.42 & $3.22_{\pm 0.27}$ & $2.21_{\pm 0.24}$ & 1.01 \\
& RePaint & $4.75_{\pm 1.36}$ & $3.01_{\pm 1.26}$ & 1.74 & $4.76_{\pm 1.04}$ & $3.84_{\pm 1.23}$ & 0.92 & $5.96_{\pm 0.27}$ & $5.74_{\pm 0.21}$ & 0.22 \\

\noalign{\vskip 0.3ex}\cdashline{3-11}\noalign{\vskip 0.3ex}
\multirow{2}{*}{\rotatebox{90}{\bf \small NJ}}
& Flow-R & $2.78_{\pm 1.27}$ & $1.83_{\pm 1.41}$ & 0.95 & $-0.59_{\pm 1.85}$ & $-3.58_{\pm 2.08}$ & 2.98 & $7.53_{\pm 0.39}$ & $6.44_{\pm 0.44}$ & 1.08 \\
& RePaint & $2.46_{\pm 1.14}$ & $1.86_{\pm 1.08}$ & 0.60 & $-2.95_{\pm 1.77}$ & $-3.96_{\pm 2.33}$ & 1.01 & $9.36_{\pm 0.34}$ & $8.93_{\pm 0.41}$ & 0.43 \\

\noalign{\vskip 0.3ex}\cdashline{3-11}\noalign{\vskip 0.3ex}

\multirow{2}{*}{\rotatebox{90}{\bf \small OS}}
& Flow-R & $2.97_{\pm 0.44}$ & $1.52_{\pm 0.44}$ & 1.45 & $-1.20_{\pm 1.72}$ & $-5.25_{\pm 1.76}$ & 4.05 & $7.90_{\pm 0.41}$ & $7.02_{\pm 0.39}$ & 0.88 \\
& RePaint & $3.52_{\pm 0.36}$ & $1.82_{\pm 0.25}$ & 1.70 & $-1.21_{\pm 2.14}$ & $-5.03_{\pm 1.70}$ & 3.82 & $10.62_{\pm 0.43}$ & $10.09_{\pm 0.41}$ & 0.53 \\

\bottomrule[0.4ex]
\end{tabular}
}
\end{table*}

\subsection{Qualitative Evaluation of {\name}}

\begin{figure*}[htb!]
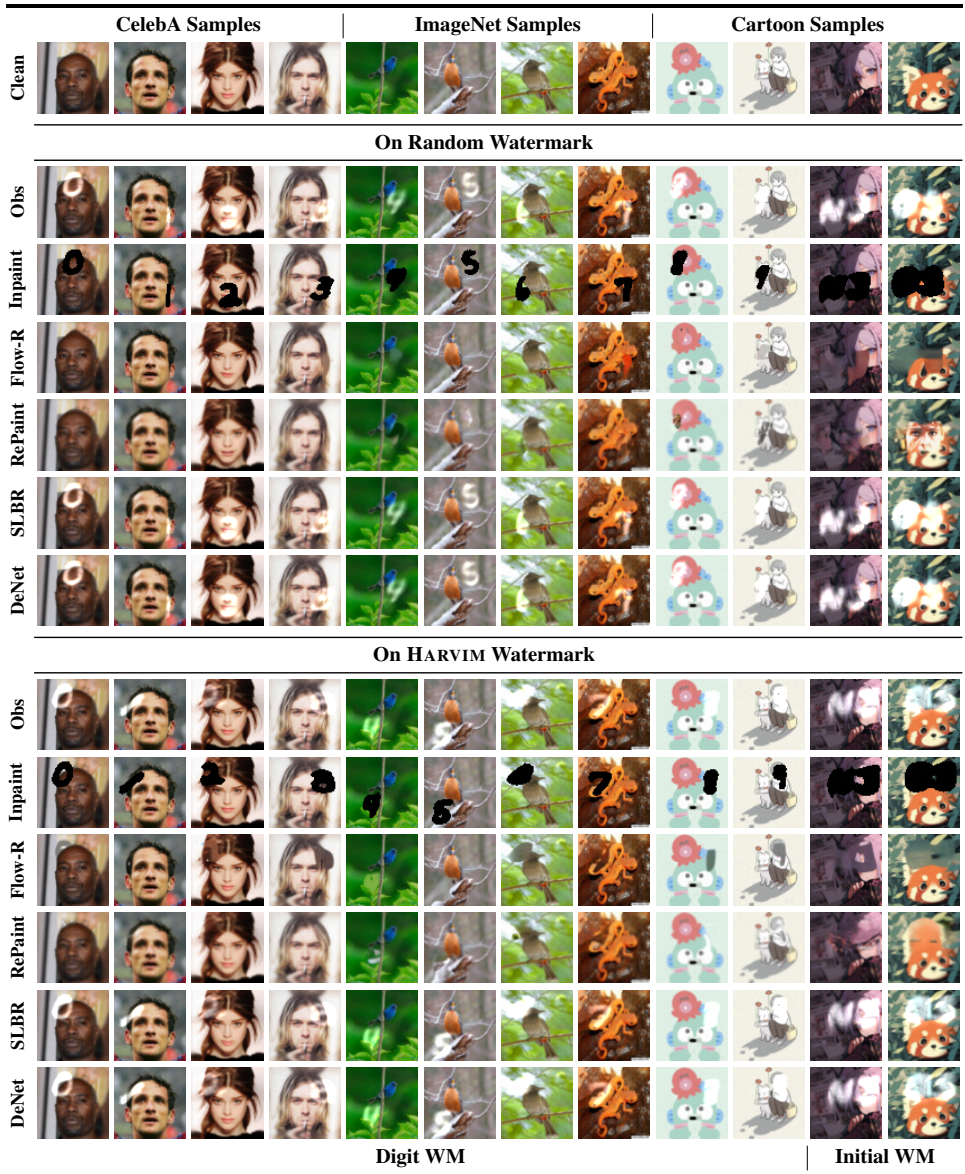

\centering
\resizebox{0.75\textwidth}{!}{
\renewcommand{\tabcolsep}{1.5pt}
\def\figwidth{0.07\linewidth}%
\newcommand{\authornote}[1]{
\adjustbox{rotate=90}{\parbox{\figwidth}{\small \bf \centering #1}}
}

\begin{tabular}{*{15}{c}}


\toprule[0.4ex]
& \multicolumn{4}{c}{\bf CelebA Samples} \vline
& \multicolumn{4}{c}{\bf ImageNet Samples} \vline
& \multicolumn{4}{c}{\bf Cartoon Samples} \\



\noalign{\vskip 0.5ex}

\authornote{Clean}     &  
\loadimage{figures/main}{width=\figwidth}{0}{12}\\
\cmidrule[0.15ex]{2-13}
\multicolumn{13}{c}{\bf On Random Watermark}  \\
\cmidrule[0.15ex]{2-13}

\authornote{Obs}     &  
\loadimage{figures/main}{width=\figwidth}{12}{12}\\
\authornote{Inpaint}     &  
\loadimage{figures/main}{width=\figwidth}{24}{12}\\
\authornote{Flow-R}     &  
\loadimage{figures/main}{width=\figwidth}{36}{12}\\
\authornote{RePaint}     &  
\loadimage{figures/main}{width=\figwidth}{48}{12}\\
\authornote{SLBR}     &  
\loadimage{figures/main}{width=\figwidth}{60}{12}\\
\authornote{DeNet}     &  
\loadimage{figures/main}{width=\figwidth}{72}{12}\\

\cmidrule[0.15ex]{2-13}
\multicolumn{13}{c}{\bf On {\name} Watermark}  \\
\cmidrule[0.15ex]{2-13}

\authornote{Obs}     &  
\loadimage{figures/main}{width=\figwidth}{96}{12}\\
\authornote{Inpaint}     &  
\loadimage{figures/main}{width=\figwidth}{108}{12}\\
\authornote{Flow-R}     &  
\loadimage{figures/main}{width=\figwidth}{120}{12}\\
\authornote{RePaint}     &  
\loadimage{figures/main}{width=\figwidth}{132}{12}\\
\authornote{SLBR}     &  
\loadimage{figures/main}{width=\figwidth}{144}{12}\\
\authornote{DeNet}     &  
\loadimage{figures/main}{width=\figwidth}{156}{12}\\


& \multicolumn{10}{c}{\bf Digit WM} \vline
& \multicolumn{2}{c}{\bf Initial WM} \\


\bottomrule[0.4ex]

\end{tabular}
}
\caption{
Watermark removal performance of \textit{worse-case} Flow-R and RePaint, and \textit{blind-case} SLBR and DeNet. 
``Obs''and ``Inpaint'' show watermarked and surrogate inpainting images respectively.  
}
\label{fig:main}
\end{figure*}

We conclude this section by providing careful analysis on how {\name} learns watermarks in order to make them hard-to-remove. We visualize
reconstructions generated using different methods applied to random and {\name} learned watermarks. Results are shown in Fig \ref{fig:main}. 
Due to space constraints, one sample is presented for each watermark.	

From Fig \ref{fig:main}, {\name} increased the difficulty of watermark removal while simultaneously preserving both image and watermark readability.
To achieve this, it selected regions with abundant visual details as {hard-to-reconstruct regions} to place watermarks. 
Importantly, we found these details are likely overlooked even by human readers.
For example, in three out of four CelebA images, {\name} placed watermarks along the boundaries between human hair and the background. 
These placements caused both Flow-R and RePaint to fail in accurately reconstructing the textures. 
Similarly, watermarks were put on leafy backgrounds on ImageNet images, leading to further failures of the two models. 

Interestingly, in column 8, both model failed to reconstruct the smaller lizard masked by the digit logo watermark. Furthermore, they both misinterpreted this lizard as part of the larger one. 
Given that Flow-R and RePaint employed generative priors of different architectures trained on distinct datasets, 
this ``coincidence'' can be considered as concrete evidence that the \textit{hard-to-reconstruct region} is an intrinsic characteristic of the image, learned by different generative priors trained in diverse scenes. 
By targeting this intrinsic characteristic, {\name} shows strong transferability.

\ExplSyntaxOn
\cs_set:Npn \loadimage #1#2#3#4 {
    \seq_clear:N \l_tmpa_seq
    \int_step_inline:nnn {#3} {#3 + #4 - 1} {
        \seq_put_right:Nx \l_tmpa_seq {
            \exp_not:N \includegraphics
                [\exp_not:n {#2}]
                {#1/##1.png}
        }
    }
    \seq_use:Nn \l_tmpa_seq {&}
}

\cs_set:Npn \calctotalwidth #1#2 {
    \fp_eval:n {
        (#2) * (#1) + (#2 - 1) * (\tabcolsep)
    } pt
}
\ExplSyntaxOff

\section{Related Works}
\label{sec:related}

\textbf{Copyright Protection.}
DGM-based AI tools have raised concerns about 
unauthorized use of copyrighted images, including style transfer~\citep{kim2022diffusionclip}, 
personalization~\citep{gal2022image,ruiz2023dreambooth}, and image editing~\citep{choi2023custom,shi2024dragdiffusion}. 
Recent studies framed data protection as a problem of \textit{adversarial attacks}.
By introducing imperceptible perturbations to protected images, these approaches aim to degrade AI performance on the affected data~\citep{salman2023raising,liang2021visible,van2023anti}. 
Particularly,
PhotoGuard~\citep{salman2023raising} attacked a text-to-image model by perturbing its latent code, aligning generations with an unrelated dummy image. 
Glaze~\citep{shan2023glaze} further employed a style-transfer model to minimize the similarity of generated images to the protected content. 
AdvDM~\citep{liang2023adversarial} targeted on diffusion-based models by minimizing the likelihood of perturbed images; and \cite{liang2023mist} added a texture-targeting loss for improved robustness. 
To defend against personalized DreamBooth~\citep{ruiz2023dreambooth}, 
\cite{van2023anti} learned perturbations to degrade its training performance using a bi-level optimization framework, 
with an approximate solution proposed by neglecting the trajectories in the lower-level optimization.
\cite{liu2024metacloak} improved this process by using meta-learning to attack an ensemble of models. 
However, existing methods specifically targeted DGMs that cause the misuse, 
and their attack-based solutions are highly specialized, making generalization challenging~\citep{demontis2019adversarial}.
In contrast, our {\name} identifies a hard-to-reconstruct region of the image and places a visible watermark, rendering the image unusable. In this way, {\name} provides protection agnostic to misuse scenarios.


\textbf{Visible Watermarking and Removal.}
Visible watermarks have been widely used to prevent piracy \citep{braudaway1996protecting,cox1997secure,mohanty1999dual,kankanhalli1999adaptive}. 
Early works resorted to signal processing technique to enhance the robustness of watermark \citep{podilchuk1998image,kankanhalli1999adaptive,hu2001wavelet}.
In response, watermark removal has also accumulated a vast literature~\citep{dekel2017effectiveness,cheng2018large,leng2024removing}. 
When the watermark location is known, inverse-problem-based solvers can provide strong reconstructions~\citep{khachaturov2021markpainting}, as also verified in our experiments. 
However, these methods are ineffective when location information is unavailable, and obtaining human 
labeling is often impractical~\citep{liu2022watermark}.
The advance of deep learning further stimulated the end-to-end blind watermark removal models. 
Early works used image translation methods to generate clean images from watermarked observations in a single step~\citep{cao2019generative,li2019towards}. 
Later, \cite{cun2021split,liang2021visible,liu2022watermark} separated the processes of locating and removing watermarks into two distinct steps, achieving more effective results.
These methods have posed remarkable performance on removing visible watermarking \citep{dekel2017effectiveness}. 
However, the primary focus of watermarking in the AI era has shifted to attack-based protection and invisible watermarking on AI-generated contents, leaving robust visible watermarking unsolved. 
In this work, we proposed a new learning-based visible watermarking and experimented with both inverse problem solver and two-stage blind watermark removal methods. 
Empirically, {\name} learns stronger watermarks to defeat all these methods.


\textbf{Watermarking in AI Era.}
The advance of vision and language foundation models~\citep{radford2019language,dhariwal2021diffusion,wei2022chain,rombach2022high,bordes2024introduction,liu2024toward,zhu2024sora} have raised ethical concerns about the potential misuse of AI-generated content (AIGC), such as deepfake~\citep{westerlund2019emergence}, plagiarism~\citep{kirchenbauer2023watermark,lau2024protecting}, and others~\citep{guo2023aigc}. 
To address these challenges, 
invisible watermarking have been proposed for embedding in the output data \citep{zhao2023recipe,liu2024survey,karki2024deep}. 
These watermarks do not affect normal use of AIGC,
but in case of misuse such as fake news, they can be extracted to trace the source of the generated content. 
For example, in watermarking vision models, an encoder and decoder are trained to generate and extract watermarks, and the vision model is often fine-tuned jointly to avoid performance degradation~\citep{yu2020responsible,fernandez2023stable,mareen2024blind,an2024benchmarking}.
For language models, the generation process is altered by increasing the likelihood of certain words while decreasing that of others. This creates a traceable pattern in the generated texts~\citep{kirchenbauer2023watermark,liu2024survey}. 
Notably, these watermarking techniques have objectives orthogonal to ours. 
They aim to ensure the traceability of AIGC, while ours targets to protect copyrighted contents created by human artists.  

\section{Conclusion}
\label{sec:conclusion}

In this paper, we introduce {\name}, a new copyright protection paradigm in the AI era. 
Unlike existing attack-based approaches, {\name} requires no domain knowledge of misuse scenarios or mechanisms.
{\name} bridges inverse problems and copyright protection and formulates learning-based \textit{hard-to-remove} visible watermarking as a bi-level optimization problem. 
Built upon recent optimality guarantees for inverse problems, 
we propose a new meta-learning solution for {\name}. 
We validate the effectiveness of our algorithm across watermarking scenarios.
Encouraged by the promising results, we identify two exciting directions for future work.
First, we will explore a training-free version of Harvim that embeds open-ended text watermarks, leveraging recent advances in image editing to enhance real-time efficiency.
Second, we will investigate the theoretical guarantees of the Harvim framework from two perspectives:
(1) determining the minimum distortion (masking) required to ensure a watermark is provably unremovable, and
(2) characterizing the maximum tolerable noise or perturbation under which a personalized concept can still be provably learned using DreamBooth.



\section*{Acknowledgment}
This material is based upon work supported by the U.S. Department of Energy, Office of Science Energy Earthshot Initiative as part of the project ``Learning reduced models under extreme data conditions for design and rapid decision-making in complex systems'' under Award \#DE-SC0024721.


\bibliography{ref}

\newpage

\onecolumn

\appendix
\clearpage

\title{Appendix}
\onecolumn

\section{Implementation Details}
\label{app:details}

In this section we provide more technical details about our implementation of watermark generators. 
Hyper-parameters and computing infrastructure information is also provided.

\subsection{Watermark Generator}
\label{app:wm-details}

In this work, we use MNIST digits~\citep{lecun1998gradient} and MNIST letters~\citep{cohen2017emnist} to create watermarks. 
Generated watermarks are grayscaled of size $ 64 \times 64$.
Watermark generators are parameterized by a three-hidden-layer MLP conditional VAEs following \citet{sohn2015learning}. 

\textbf{Controllable Location.}
To make watermark location controllable, when training the watermark generator, 
we put digits and letters in various locations in the image, by assigning different padding sizes to four sides. 
Based on this, we calculated the left and bottom padding ratios lying in $[0, 1]$. 
To be more specific, the watermark is placed at the \textit{leftmost} region when the left padding ratio takes 0, and at the \textit{rightmost} region when it takes 1. 
Bottom padding ratio functions in a similar way. 
In the training, the two padding ratios, together with the digit index (0-9), are used as conditions.

\textbf{Watermark Optimization.}
In Alg \ref{alg:main}, {\name} uses the fixed pre-trained CVAE and optimizes its latent code and two padding ratios to learn watermarks. 
This parameterization allows us to maintain a good \textbf{watermark readability}. 
To further satisfies the image readability, we define $\mathcal R (\mv) = \| \mv \|_1$ to avoid learning an excessive watermark.

\textbf{Differentiable Approximation for Mask Matrix.}
To obtain a differentiable approximation for masking matrix in inpainting, we define 
\begin{align*}
    \Amat_m = \text{diag}\left(\text{sig}\left(\frac{m_1 - \alpha}{\beta}\right), \dots, \text{sig}\left(\frac{m_n - \alpha}{\beta}\right)  \right),
\end{align*}
and use $\alpha=0.15, \beta=0.01$. 
See inpainting observations in Figure \ref{fig:main} generated with this differentiable approximation.

\textbf{Initialization}
Due to the highly non-smooth nature in location parameters, 
in {\name}, we conduct a 3-by-3 grid search on watermark locations before running the complete algorithm. This search is based on the reconstruction quality from a 50-step MLE solution, as detailed in \cite{liu2023aipo}.

\subsection{Hyperparameters}

We summarize all hyperparameters used in this paper in Table \ref{tab:hparams}, which we found working well. 
In execution, we rescale $\mathcal{R}(\mv)$ terms before tuning its weight to avoid bearing with the magnitude difference. 
Here we also provide a few clarification on some choices.

First, the
``coefficient of watermark regularizer'' in Eq \eqref{eq:inpaint-mask} (i.e., 
norm of the watermark), was tuned by monitoring the watermark size. 
We note that a strong regularization will make the watermark disappear, and a weak one will let the watermark cover the whole image. We chose 0.001 as it allowed watermark size to remain stable, i.e., close to the initial size during the optimization process. Note that it was not tuned based on watermark removal performance.

Second, the two ``smoothing factors'' $\alpha, \beta$, as discussed above, were not treated as tunable hyperparameter to benefit {\name} performance. 
Instead, we chose the two solely to make the sigmoid function have a fairly wide range $[0, 1]$ when its input (pixel) lies in $[0, 1]$. 

\begin{table}[htbp!]
\centering
\caption{
Hyper-parameters of different methods.
}
\label{tab:hparams}
\renewcommand{\tabcolsep}{4pt}
\resizebox{0.7\linewidth}{!}{
\begin{tabular}{r r cc}
\toprule[0.3ex]
& & Digit Logo Watermarks & Initial Watermarks \\

\cmidrule{3-4}
                                    & HParam          & Value   &         Value \\
\cmidrule[0.3ex]{2-4} 
\multirow{7}{*}{{\name}}            & Learning Rate     & 0.05    & 0.05          \\
                                    & Optimizer & Default AdamW \citep{ilya2018decoupled} & Default AdamW \citep{ilya2018decoupled} \\
                                    & Coeff. of $\mathcal{R}(\mv)$ (Eq \eqref{eq:inpaint-mask}) & 0.001 & 0.001 \\
                                    & $\alpha, \beta$ (Eq \eqref{eq:inpaint-mask}) & $\alpha=0.15, \beta=0.01$ & $\alpha=0.15, \beta=0.01$ \\
                                    & Meta Step $K$       & 1   & 1 \\
                                    & Targeted $\lambda$  & 1   & 1 \\
                                    & Step Size for $\lambda$  & \multicolumn{2}{c}{Using dynamic strategies from \citet{liu2023aipo}.}\\

\cmidrule{2-4} 
\multirow{2}{*}{{Flow-R}}        & $\lambda$    & 1   & 1 \\
                                    & Others  & \multicolumn{2}{c}{Identical to \citet{liu2023aipo}} \\
\cmidrule{2-4} 
\multirow{2}{*}{{Repaint}}          & Batch Size & 10 & 10 \\ 
                                    & Others  & \multicolumn{2}{c}{Identical to \citet{lugmayr2022repaint}} \\
\cmidrule{2-4} 
\multirow{1}{*}{{SLBR}}             & All  & \multicolumn{2}{c}{Identical to \citet{liang2021visible}} \\
\multirow{1}{*}{{DeNet}}             & All  & \multicolumn{2}{c}{Identical to \citet{sun2023denet}} \\

\bottomrule[0.3ex]
\end{tabular}
}
\end{table}

\subsection{Computing Infrastructure}
\label{app:infra}

Our experiments were conducted on a NVIDIA A6000 48GB GPU.
Our watermark generators were trained on a CPU-only machine.

\subsection{Testing Datasets}
\label{app:data}
We tested {\name} on 100 validation samples from CelebA and ImageNet datasets respectively. 
More specifically, to guarantee reproducibility, we used the first 100 CelebA validation samples from \cite{whang21solve}; and the first 100 ImageNet samples from public subset on \url{https://github.com/EliSchwartz/imagenet-sample-images}.

\section{More Experiment Results}
\label{app:results}

In this section we provide more experimental results. 
In particular, Table \ref{tab:quant-full} reports the complete version of Table \ref{tab:quant}. 
SLBR failed to recognize the two types of watermark, as indicated by its low ``Before'' and ``After'' values: these values, as detailed in Sec \ref{sec:experiment}, refer to how much reconstruction $\Tilde{\xv}$ is better than observations $\yv$, i.e., $\text{PSNR}(\Tilde{\xv}, \xv_T) - \text{PSNR}(\yv, \xv_T)$. 
See Figure \ref{fig:main} for concrete examples of SLBR failures.

\begin{table*}[htb!]
\definecolor{verylightgray}{gray}{0.95}
\setlength{\tabcolsep}{4pt}  

\centering
\caption{
(Complete version of Table \ref{tab:quant})
Performance of different watermark removal methods on random and {\name} watermarks.
The performance is evaluated based on how the reconstruction is better than the observation in terms of PSNR, SSIM, and LPIPS respectively. 
}
\label{tab:quant-full}

\resizebox{0.8\linewidth}{!}{%
\begin{tabular}{
c >{\bfseries}r 
cc>{\columncolor{verylightgray}}c 
cc>{\columncolor{verylightgray}}c 
cc>{\columncolor{verylightgray}}c
}
\toprule[0.4ex]
\multicolumn{11}{c}{\bf CelebA } \\
\cmidrule[0.15ex]{2-11}

& & \multicolumn{3}{c}{\bf PSNR} & \multicolumn{3}{c}{\bf SSIM $\times$ 100} & \multicolumn{3}{c}{\bf LPIPS $\times$ 100} \\
\cmidrule[0.15ex]{3-11}
& & \bf Random & \bf {\name} & \bf Imp ($\uparrow$) & \bf Random & \bf {\name} & \bf Imp ($\uparrow$) & \bf Random & \bf {\name} & \bf Imp ($\uparrow$) \\
\cmidrule(lr){3-5} \cmidrule(lr){6-8} \cmidrule(lr){9-11}

\multirow{4}{*}{\rotatebox{90}{\bf \small DIG}} 
& Flow-R & $13.02_{\pm 0.37}$ & $7.57_{\pm 0.66}$ & 5.44 & $6.71_{\pm 0.24}$ & $3.82_{\pm 0.40}$ & 2.89 & $3.22_{\pm 0.27}$ & $2.21_{\pm 0.24}$ & 1.01 \\
& RePaint & $13.45_{\pm 0.41}$ & $11.57_{\pm 0.46}$ & 1.89 & $9.63_{\pm 0.25}$ & $8.62_{\pm 0.33}$ & 1.01 & $5.96_{\pm 0.27}$ & $5.74_{\pm 0.21}$ & 0.22 \\
& SLBR & $0.10_{\pm 0.01}$ & $0.11_{\pm 0.02}$ & -0.01  & $0.81_{\pm 0.02}$ & $0.78_{\pm 0.02}$ & 0.03 & $0.12_{\pm 0.03}$ & $0.12_{\pm 0.05}$ & 0.00 \\
& DeNet & $0.10_{\pm 0.01}$ & $0.12_{\pm 0.02}$ & -0.02 & $0.83_{\pm 0.02}$ & $0.82_{\pm 0.02}$ & 0.01  & $0.12_{\pm 0.03}$ & $0.18_{\pm 0.04}$ & -0.06 \\

\noalign{\vskip 0.3ex}\cdashline{3-11}\noalign{\vskip 0.3ex}
\multirow{4}{*}{\rotatebox{90}{\bf \small NJ}}
& Flow-R & $9.31_{\pm 0.34}$ & $7.33_{\pm 0.36}$ & 1.98 & $11.40_{\pm 0.58}$ & $9.34_{\pm 0.61}$ & 2.06 & $7.53_{\pm 0.39}$ & $6.44_{\pm 0.44}$ & 1.08 \\
& RePaint & $10.25_{\pm 0.34}$ & $9.99_{\pm 0.38}$ & 0.26 & $14.08_{\pm 0.60}$ & $13.07_{\pm 0.67}$ & 1.01 & $9.36_{\pm 0.34}$ & $8.93_{\pm 0.41}$ & 0.43 \\
& SLBR & $0.00_{\pm 0.00}$ & $0.01_{\pm 0.00}$ & 0.00 & $0.32_{\pm 0.02}$ & $0.35_{\pm 0.02}$ & -0.03  & $-0.40_{\pm 0.04}$ & $-0.36_{\pm 0.04}$ & -0.03 \\
& DeNet & $0.00_{\pm 0.00}$ & $0.01_{\pm 0.00}$ & 0.00 & $0.36_{\pm 0.02}$ & $0.38_{\pm 0.02}$ & -0.02  & $-0.37_{\pm 0.04}$ & $-0.36_{\pm 0.04}$ & -0.01 \\

\noalign{\vskip 0.3ex}\cdashline{3-11}\noalign{\vskip 0.3ex}
\multirow{4}{*}{\rotatebox{90}{\bf \small OS}}
& Flow-R & $9.57_{\pm 0.33}$ & $7.21_{\pm 0.31}$ & 2.36 & $11.43_{\pm 0.62}$ & $8.01_{\pm 0.53}$ & 3.42 & $7.90_{\pm 0.41}$ & $7.02_{\pm 0.39}$ & 0.88 \\
& RePaint & $10.39_{\pm 0.36}$ & $9.44_{\pm 0.37}$ & 0.96 & $15.68_{\pm 0.64}$ & $12.86_{\pm 0.62}$ & 2.81 & $10.62_{\pm 0.43}$ & $10.09_{\pm 0.41}$ & 0.53 \\
& SLBR & $-0.01_{\pm 0.00}$ & $0.00_{\pm 0.00}$ & 0.00  & $0.20_{\pm 0.02}$ & $0.25_{\pm 0.02}$ & -0.05  & $-0.46_{\pm 0.05}$ & $-0.38_{\pm 0.05}$ & -0.09\\ 
& DeNet & $0.00_{\pm 0.00}$ & $0.00_{\pm 0.00}$ & 0.00  & $0.26_{\pm 0.02}$ & $0.32_{\pm 0.02}$ & -0.06  & $-0.40_{\pm 0.05}$ & $-0.33_{\pm 0.05}$ & -0.07\\

\cmidrule[0.15ex]{2-11}

\multicolumn{11}{c}{\bf ImageNet } \\
\cmidrule[0.15ex]{2-11}

& & \multicolumn{3}{c}{\bf PSNR} & \multicolumn{3}{c}{\bf SSIM $\times$ 100} & \multicolumn{3}{c}{\bf LPIPS $\times$ 100} \\
\cmidrule[0.15ex]{3-11}
& & \bf Random & \bf {\name} & \bf Imp ($\uparrow$) & \bf Random & \bf {\name} & \bf Imp ($\uparrow$) & \bf Random & \bf {\name} & \bf Imp ($\uparrow$) \\
\cmidrule(lr){3-5} \cmidrule(lr){6-8} \cmidrule(lr){9-11}

\multirow{4}{*}{\rotatebox{90}{\bf \small DIG}}
& Flow-R & $9.61_{\pm 0.45}$ & $6.09_{\pm 0.62}$ & 3.52 & $4.49_{\pm 0.35}$ & $2.72_{\pm 0.45}$ & 1.77 & $3.22_{\pm 0.27}$ & $2.21_{\pm 0.24}$ & 1.01 \\
& RePaint & $9.61_{\pm 0.48}$ & $8.56_{\pm 0.53}$ & 1.05 & $8.80_{\pm 0.39}$ & $8.40_{\pm 0.44}$ & 0.41 & $5.96_{\pm 0.27}$ & $5.74_{\pm 0.21}$ & 0.22 \\
& SLBR & $0.13_{\pm 0.01}$ & $0.12_{\pm 0.02}$ & 0.01 & $1.05_{\pm 0.04}$ & $1.01_{\pm 0.05}$ & 0.04 & $-0.36_{\pm 0.07}$ & $-0.14_{\pm 0.07}$ & -0.22 \\ 
& DeNet & $0.13_{\pm 0.01}$ & $0.14_{\pm 0.01}$ & -0.01  & $1.13_{\pm 0.04}$ & $1.12_{\pm 0.04}$ & 0.01 & $-0.31_{\pm 0.07}$ & $-0.09_{\pm 0.07}$ & -0.22 \\

\noalign{\vskip 0.3ex}\cdashline{3-11}\noalign{\vskip 0.3ex}
\multirow{4}{*}{\rotatebox{90}{\bf \small NJ}}
& Flow-R & $8.46_{\pm 0.37}$ & $7.21_{\pm 0.42}$ & 1.25 & $5.12_{\pm 0.73}$ & $3.42_{\pm 0.75}$ & 1.70 & $7.53_{\pm 0.39}$ & $6.44_{\pm 0.44}$ & 1.08 \\
& RePaint & $7.20_{\pm 0.31}$ & $6.69_{\pm 0.40}$ & 0.50 & $3.76_{\pm 0.73}$ & $2.79_{\pm 0.82}$ & 0.96 & $9.36_{\pm 0.34}$ & $8.93_{\pm 0.41}$ & 0.43 \\
& SLBR & $0.00_{\pm 0.00}$ & $0.00_{\pm 0.00}$ & 0.00  & $0.36_{\pm 0.04}$ & $0.39_{\pm 0.04}$ & -0.03 & $-1.14_{\pm 0.09}$ & $-0.96_{\pm 0.08}$ & -0.18 \\
& DeNet & $0.00_{\pm 0.00}$ & $0.00_{\pm 0.00}$ & 0.00  & $0.46_{\pm 0.04}$ & $0.47_{\pm 0.04}$ & -0.01 & $-1.04_{\pm 0.07}$ & $-0.90_{\pm 0.07}$ & -0.14 \\

\noalign{\vskip 0.3ex}\cdashline{3-11}\noalign{\vskip 0.3ex}
\multirow{4}{*}{\rotatebox{90}{\bf \small OS}}
& Flow-R & $8.58_{\pm 0.34}$ & $7.18_{\pm 0.39}$ & 1.40 & $4.28_{\pm 0.69}$ & $1.82_{\pm 0.69}$ & 2.46 & $7.90_{\pm 0.41}$ & $7.02_{\pm 0.39}$ & 0.88 \\
& RePaint & $7.44_{\pm 0.33}$ & $6.60_{\pm 0.38}$ & 0.84 & $4.03_{\pm 0.68}$ & $2.03_{\pm 0.77}$ & 2.00 & $10.62_{\pm 0.43}$ & $10.09_{\pm 0.41}$ & 0.53 \\
& SLBR & $-0.01_{\pm 0.00}$ & $-0.01_{\pm 0.00}$ & 0.00  & $0.24_{\pm 0.04}$ & $0.31_{\pm 0.04}$ & -0.07 & $-1.13_{\pm 0.08}$ & $-0.95_{\pm 0.08}$ & -0.17 \\
& DeNet & $0.00_{\pm 0.00}$ & $0.00_{\pm 0.00}$ & 0.00  & $0.34_{\pm 0.04}$ & $0.41_{\pm 0.04}$ & -0.07 & $-1.03_{\pm 0.07}$ & $-0.82_{\pm 0.07}$ & -0.21 \\

\cmidrule[0.15ex]{2-11}
\multicolumn{11}{c}{\bf Cartoon} \\
\cmidrule[0.15ex]{2-11}

& & \multicolumn{3}{c}{\bf PSNR} & \multicolumn{3}{c}{\bf SSIM $\times$ 100} & \multicolumn{3}{c}{\bf LPIPS $\times$ 100} \\
\cmidrule[0.15ex]{3-11}
& & \bf Random & \bf {\name} & \bf Imp ($\uparrow$) & \bf Random & \bf {\name} & \bf Imp ($\uparrow$) & \bf Random & \bf {\name} & \bf Imp ($\uparrow$) \\
\cmidrule(lr){3-5} \cmidrule(lr){6-8} \cmidrule(lr){9-11}

\multirow{4}{*}{\rotatebox{90}{\bf \small DIG}}
& Flow-R & $6.53_{\pm 1.18}$ & $-0.86_{\pm 1.74}$ & 7.39 & $2.55_{\pm 0.77}$ & $-0.88_{\pm 1.25}$ & 3.42 & $3.22_{\pm 0.27}$ & $2.21_{\pm 0.24}$ & 1.01 \\
& RePaint & $4.75_{\pm 1.36}$ & $3.01_{\pm 1.26}$ & 1.74 & $4.76_{\pm 1.04}$ & $3.84_{\pm 1.23}$ & 0.92 & $5.96_{\pm 0.27}$ & $5.74_{\pm 0.21}$ & 0.22 \\
& SLBR & $0.12_{\pm 0.05}$ & $0.14_{\pm 0.05}$ & -0.02  & $1.15_{\pm 0.17}$ & $0.99_{\pm 0.10}$ & 0.15  & $0.26_{\pm 0.14}$ & $0.49_{\pm 0.13}$ & -0.23 \\
& DeNet & $0.16_{\pm 0.03}$ & $0.14_{\pm 0.05}$ & 0.03 & $1.09_{\pm 0.11}$ & $0.98_{\pm 0.09}$ & 0.11 & $0.27_{\pm 0.14}$ & $0.45_{\pm 0.14}$ & -0.19 \\

\noalign{\vskip 0.3ex}\cdashline{3-11}\noalign{\vskip 0.3ex}
\multirow{4}{*}{\rotatebox{90}{\bf \small NJ}}
& Flow-R & $2.78_{\pm 1.27}$ & $1.83_{\pm 1.41}$ & 0.95 & $-0.59_{\pm 1.85}$ & $-3.58_{\pm 2.08}$ & 2.98 & $7.53_{\pm 0.39}$ & $6.44_{\pm 0.44}$ & 1.08 \\
& RePaint & $2.46_{\pm 1.14}$ & $1.86_{\pm 1.08}$ & 0.60 & $-2.95_{\pm 1.77}$ & $-3.96_{\pm 2.33}$ & 1.01 & $9.36_{\pm 0.34}$ & $8.93_{\pm 0.41}$ & 0.43 \\
& SLBR & $-0.02_{\pm 0.01}$ & $-0.06_{\pm 0.03}$ & 0.04  & $0.38_{\pm 0.09}$ & $0.21_{\pm 0.17}$ & 0.17 & $-0.32_{\pm 0.24}$ & $-0.37_{\pm 0.29}$ & 0.04 \\
& DeNet & $-0.01_{\pm 0.00}$ & $-0.03_{\pm 0.01}$ & 0.02  & $0.45_{\pm 0.10}$ & $0.42_{\pm 0.10}$ & 0.02 & $-0.29_{\pm 0.23}$ & $-0.19_{\pm 0.19}$ & -0.09 \\

\noalign{\vskip 0.3ex}\cdashline{3-11}\noalign{\vskip 0.3ex}
\multirow{4}{*}{\rotatebox{90}{\bf \small OS}}
& Flow-R & $2.97_{\pm 0.44}$ & $1.52_{\pm 0.44}$ & 1.45 & $-1.20_{\pm 1.72}$ & $-5.25_{\pm 1.76}$ & 4.05 & $7.90_{\pm 0.41}$ & $7.02_{\pm 0.39}$ & 0.88 \\
& RePaint & $3.52_{\pm 0.36}$ & $1.82_{\pm 0.25}$ & 1.70 & $-1.21_{\pm 2.14}$ & $-5.03_{\pm 1.70}$ & 3.82 & $10.62_{\pm 0.43}$ & $10.09_{\pm 0.41}$ & 0.53 \\
& SLBR & $-0.02_{\pm 0.00}$ & $-0.04_{\pm 0.01}$ & 0.02  & $0.32_{\pm 0.12}$ & $0.28_{\pm 0.11}$ & 0.03 & $-0.30_{\pm 0.27}$ & $-0.37_{\pm 0.21}$ & 0.08 \\
& DeNet & $-0.02_{\pm 0.00}$ & $-0.04_{\pm 0.01}$ & 0.02  & $0.32_{\pm 0.12}$ & $0.29_{\pm 0.11}$ & 0.03 & $-0.30_{\pm 0.27}$ & $-0.36_{\pm 0.22}$ & 0.07 \\

\bottomrule[0.4ex]
\end{tabular}
}
\end{table*}

\begin{table*}[htb!]
\definecolor{verylightgray}{gray}{0.9}

\centering
\caption{
Reconstruction quality of different watermark removal methods on random and {\name}'s learned watermarks.
Lower PSNR, SSIM, and higher LPIPS indicates better robustness against removal. 
}
\label{tab:raw}

\resizebox{0.75\linewidth}{!}{%
\begin{tabular}{
c >{\bfseries}r 
cc
cc
cc
}
\toprule[0.4ex]
\multicolumn{8}{c}{\bf CelebA } \\
\cmidrule[0.15ex]{2-8}

& & \multicolumn{2}{c}{\bf PSNR} & \multicolumn{2}{c}{\bf SSIM $\times$ 100} & \multicolumn{2}{c}{\bf LPIPS $\times$ 100} \\
\cmidrule[0.15ex]{3-8}
& & \bf Random & \bf {\name} & \bf Random & \bf {\name} & \bf Random & \bf {\name} \\
\cmidrule(lr){3-4} \cmidrule(lr){5-6} \cmidrule(lr){7-8}

\multirow{4}{*}{\rotatebox{90}{\bf \small DIG}} 
& Flow-R & $35.15_{\pm 0.36}$ & $30.32_{\pm 0.59}$ & $96.67_{\pm 0.18}$ & $94.02_{\pm 0.33}$ & $0.57_{\pm 0.09}$ & $1.36_{\pm 0.14}$
\\
& RePaint & $35.34_{\pm 0.42}$ & $33.99_{\pm 0.48}$ & $97.38_{\pm 0.15}$ & $96.67_{\pm 0.23}$ & $0.64_{\pm 0.10}$ & $0.69_{\pm 0.08}$
\\
& SLBR & $22.20_{\pm 0.19}$ & $22.81_{\pm 0.29}$ & $90.39_{\pm 0.24}$ & $90.64_{\pm 0.30}$ & $3.94_{\pm 0.27}$ & $3.73_{\pm 0.22}$
\\
& DeNet & $22.19_{\pm 0.19}$ & $22.82_{\pm 0.29}$ & $90.41_{\pm 0.24}$ & $90.67_{\pm 0.30}$ & $3.94_{\pm 0.28}$ & $3.67_{\pm 0.21}$
\\

\noalign{\vskip 0.3ex}\cdashline{3-8}\noalign{\vskip 0.3ex}
\multirow{4}{*}{\rotatebox{90}{\bf \small NJ}}
& Flow-R  & $23.51_{\pm 0.33}$ & $21.58_{\pm 0.29}$ & $83.84_{\pm 0.50}$ & $82.46_{\pm 0.49}$ & $4.33_{\pm 0.22}$ & $4.89_{\pm 0.23}$
\\
& RePaint & $24.45_{\pm 0.35}$ & $24.23_{\pm 0.35}$ & $84.83_{\pm 0.55}$ & $84.41_{\pm 0.56}$ & $4.09_{\pm 0.20}$ & $4.00_{\pm 0.27}$
\\
& SLBR  & $14.21_{\pm 0.11}$ & $14.26_{\pm 0.16}$ & $72.47_{\pm 0.36}$ & $73.16_{\pm 0.48}$ & $12.18_{\pm 0.36}$ & $11.63_{\pm 0.39}$
\\
& DeNet  & $14.22_{\pm 0.11}$ & $14.26_{\pm 0.16}$ & $72.52_{\pm 0.36}$ & $73.19_{\pm 0.48}$ & $12.16_{\pm 0.37}$ & $11.62_{\pm 0.39}$
\\

\noalign{\vskip 0.3ex}\cdashline{3-8}\noalign{\vskip 0.3ex}
\multirow{4}{*}{\rotatebox{90}{\bf \small OS}}
& Flow-R  & $22.44_{\pm 0.31}$ & $20.34_{\pm 0.25}$ & $79.54_{\pm 0.53}$ & $78.48_{\pm 0.48}$ & $6.36_{\pm 0.30}$ & $6.51_{\pm 0.25}$
\\
& RePaint & $23.27_{\pm 0.37}$ & $22.57_{\pm 0.37}$ & $82.33_{\pm 0.56}$ & $81.87_{\pm 0.57}$ & $5.12_{\pm 0.31}$ & $5.09_{\pm 0.31}$
\\
& SLBR    & $12.87_{\pm 0.11}$ & $13.13_{\pm 0.15}$ & $68.08_{\pm 0.35}$ & $70.52_{\pm 0.46}$ & $14.66_{\pm 0.37}$ & $13.89_{\pm 0.35}$
\\
& DENET   & $12.87_{\pm 0.11}$ & $13.13_{\pm 0.15}$ & $68.14_{\pm 0.36}$ & $70.58_{\pm 0.46}$ & $14.59_{\pm 0.37}$ & $13.84_{\pm 0.35}$
\\

\cmidrule[0.15ex]{2-8}

\multicolumn{8}{c}{\bf ImageNet } \\
\cmidrule[0.15ex]{2-8}

& & \multicolumn{2}{c}{\bf PSNR} & \multicolumn{2}{c}{\bf SSIM $\times$ 100} & \multicolumn{2}{c}{\bf LPIPS $\times$ 100} \\
\cmidrule[0.15ex]{3-8}
& & \bf Random & \bf {\name} & \bf Random & \bf {\name} & \bf Random & \bf {\name} \\
\cmidrule(lr){3-4} \cmidrule(lr){5-6} \cmidrule(lr){7-8}

\multirow{4}{*}{\rotatebox{90}{\bf \small DIG}}
& Flow-R  & $31.33_{\pm 0.43}$ & $28.25_{\pm 0.52}$ & $93.61_{\pm 0.26}$ & $91.73_{\pm 0.32}$ & $2.24_{\pm 0.19}$ & $3.66_{\pm 0.27}$
\\
& RePaint & $30.93_{\pm 0.49}$ & $30.23_{\pm 0.50}$ & $93.59_{\pm 0.26}$ & $93.11_{\pm 0.29}$ & $2.55_{\pm 0.22}$ & $2.93_{\pm 0.26}$
\\
& SLBR    & $21.78_{\pm 0.20}$ & $22.20_{\pm 0.25}$ & $89.44_{\pm 0.35}$ & $89.32_{\pm 0.38}$ & $8.49_{\pm 0.60}$ & $7.71_{\pm 0.58}$
\\
& DENET  & $21.79_{\pm 0.20}$ & $22.22_{\pm 0.26}$ & $89.52_{\pm 0.34}$ & $89.43_{\pm 0.37}$ & $8.44_{\pm 0.60}$ & $7.65_{\pm 0.58}$
\\

\noalign{\vskip 0.3ex}\cdashline{3-8}\noalign{\vskip 0.3ex}
\multirow{4}{*}{\rotatebox{90}{\bf \small NJ}}
& Flow-R  & $22.15_{\pm 0.34}$ & $21.14_{\pm 0.34}$ & $75.34_{\pm 0.50}$ & $74.44_{\pm 0.46}$ & $12.32_{\pm 0.49}$ & $11.99_{\pm 0.42}$
\\
& RePaint & $20.87_{\pm 0.28}$ & $20.60_{\pm 0.35}$  & $70.52_{\pm 0.47}$ & $70.39_{\pm 0.56}$ & $14.98_{\pm 0.51}$ & $14.68_{\pm 0.49}$
\\
& SLBR  & $13.70_{\pm 0.12}$ & $13.94_{\pm 0.15}$ & $69.97_{\pm 0.49}$ & $70.83_{\pm 0.52}$ & $20.40_{\pm 0.79}$ & $19.25_{\pm 0.74}$
\\
& DENET  & $13.70_{\pm 0.12}$ & $13.94_{\pm 0.15}$ & $70.07_{\pm 0.49}$ & $70.91_{\pm 0.52}$ & $20.30_{\pm 0.78}$ & $19.19_{\pm 0.74}$
\\

\noalign{\vskip 0.3ex}\cdashline{3-8}\noalign{\vskip 0.3ex}
\multirow{4}{*}{\rotatebox{90}{\bf \small OS}}
& Flow-R & $20.91_{\pm 0.31}$ & $19.89_{\pm 0.32}$ & $70.11_{\pm 0.47}$ & $69.82_{\pm 0.45}$ & $16.60_{\pm 0.58}$ & $16.65_{\pm 0.54}$
\\
& RePaint & $19.77_{\pm 0.30}$ & $19.31_{\pm 0.31}$ & $66.94_{\pm 0.43}$ & $67.11_{\pm 0.50}$ & $17.33_{\pm 0.59}$ & $17.24_{\pm 0.58}$
\\
& SLBR & $12.33_{\pm 0.12}$ & $12.71_{\pm 0.16}$ & $65.60_{\pm 0.48}$ & $67.87_{\pm 0.56}$ & $23.16_{\pm 0.77}$ & $21.50_{\pm 0.77}$
\\
& DENET & $12.34_{\pm 0.12}$ & $12.72_{\pm 0.16}$ & $65.70_{\pm 0.48}$ & $67.97_{\pm 0.56}$ & $23.07_{\pm 0.77}$ & $21.37_{\pm 0.77}$
\\

\cmidrule[0.15ex]{2-8}
\multicolumn{8}{c}{\bf Cartoon } \\
\cmidrule[0.15ex]{2-8}

& & \multicolumn{2}{c}{\bf PSNR} & \multicolumn{2}{c}{\bf SSIM $\times$ 100} & \multicolumn{2}{c}{\bf LPIPS $\times$ 100} \\
\cmidrule[0.15ex]{3-8}
& & \bf Random & \bf {\name} & \bf Random & \bf {\name} & \bf Random & \bf {\name} \\
\cmidrule(lr){3-4} \cmidrule(lr){5-6} \cmidrule(lr){7-8}

\multirow{4}{*}{\rotatebox{90}{\bf \small DIG}}
& Flow-R & $30.74_{\pm 1.31}$ & $23.73_{\pm 1.02}$& $93.89_{\pm 0.57}$ & $90.52_{\pm 0.45}$ & $1.22_{\pm 0.22}$ & $3.52_{\pm 0.40}$
\\
& RePaint & $28.30_{\pm 1.45}$ & $26.88_{\pm 0.96}$ & $93.15_{\pm 0.81}$ & $92.44_{\pm 0.84}$ & $2.60_{\pm 0.51}$ & $3.21_{\pm 0.78}$
\\
& SLBR & $24.25_{\pm 0.76}$ & $24.66_{\pm 0.84}$ & $92.03_{\pm 0.55}$ & $91.99_{\pm 1.12}$ & $2.83_{\pm 0.36}$ & $2.48_{\pm 0.34}$
\\
& DENET & $24.29_{\pm 0.77}$ & $24.65_{\pm 0.84}$ & $91.97_{\pm 0.57}$ & $91.98_{\pm 1.13}$ & $2.83_{\pm 0.36}$ & $2.52_{\pm 0.36}$
\\

\noalign{\vskip 0.3ex}\cdashline{3-8}\noalign{\vskip 0.3ex}
\multirow{4}{*}{\rotatebox{90}{\bf \small NJ}}
& Flow-R & $19.07_{\pm 0.68}$ & $18.46_{\pm 0.76}$ & $73.71_{\pm 1.16}$ & $72.86_{\pm 1.67}$ & $11.31_{\pm 1.42}$ & $11.60_{\pm 1.22}$
\\
& RePaint & $18.78_{\pm 1.05}$ & $18.53_{\pm 0.87}$ & $69.39_{\pm 0.96}$ & $70.60_{\pm 1.06}$  & $12.44_{\pm 1.10}$ & $16.02_{\pm 1.51}$
\\
& SLBR & $16.31_{\pm 0.85}$ & $16.62_{\pm 0.92}$ & $74.45_{\pm 1.07}$ & $76.46_{\pm 1.60}$ & $10.96_{\pm 1.77}$ & $10.37_{\pm 1.36}$
\\
& DENET & $16.32_{\pm 0.86}$ & $16.65_{\pm 0.93}$ & $74.52_{\pm 1.09}$ & $76.67_{\pm 1.59}$ & $10.93_{\pm 1.76}$ & $10.20_{\pm 1.30}$
\\

\noalign{\vskip 0.3ex}\cdashline{3-8}\noalign{\vskip 0.3ex}
\multirow{4}{*}{\rotatebox{90}{\bf \small OS}}
& Flow-R & $17.95_{\pm 0.71}$ & $17.08_{\pm 0.63}$ & $68.71_{\pm 0.88}$ & $68.07_{\pm 0.88}$ & $15.43_{\pm 1.10}$ & $16.66_{\pm 1.64}$
\\ 
& RePaint & $17.98_{\pm 0.81}$ & $18.32_{\pm 0.78}$ & $65.67_{\pm 1.06}$ & $67.84_{\pm 1.15}$ & $15.37_{\pm 0.71}$ & $15.39_{\pm 1.16}$
\\ 
& SLBR & $15.01_{\pm 0.93}$ & $15.58_{\pm 0.93}$ & $70.07_{\pm 1.27}$ & $73.52_{\pm 1.33}$ & $13.77_{\pm 1.73}$ & $13.04_{\pm 1.74}$
\\  
& DENET  & $15.01_{\pm 0.93}$ & $15.58_{\pm 0.93}$ & $70.07_{\pm 1.27}$ & $73.52_{\pm 1.33}$ & $13.77_{\pm 1.73}$ & $13.03_{\pm 1.74}$
\\

\bottomrule[0.4ex]
\end{tabular}
}
\end{table*}

\end{document}